\documentclass[10pt,twocolumn,letterpaper]{article}

\usepackage{wacv}              

\usepackage{graphicx}
\usepackage{amsmath}
\usepackage{amssymb}
\usepackage{booktabs}

\usepackage{url}
\usepackage{times}
\usepackage{multirow}
\usepackage{array}
\newcolumntype{?}{!{\vrule width 1.5pt}}

\usepackage[pagebackref,breaklinks,colorlinks]{hyperref}

\usepackage[capitalize]{cleveref}
\crefname{section}{Sec.}{Secs.}
\Crefname{section}{Section}{Sections}
\Crefname{table}{Table}{Tables}
\crefname{table}{Tab.}{Tabs.}

\def\wacvPaperID{8} 
\def\confName{WACV}
\def\confYear{2024}

\begin{document}

\title{On the Vulnerability of Deepfake Detectors to Attacks Generated by \\Denoising Diffusion Models}

\author{Marija Ivanovska$^{1}$, Vitomir \v{S}truc$^{1}$\\
$^{1}$Faculty of electrical engineering, University of Ljubljana\\
Tr\v{z}a\v{s}ka cesta 25, 1000 Ljubljana, Slovenia\\
{\tt\small \{marija.ivanovska, vitomir.struc\}@fe.uni-lj.si}
}
\maketitle

\begin{abstract}
The detection of malicious deepfakes is a constantly evolving problem that requires continuous monitoring of detectors to ensure they can detect image manipulations generated by the latest emerging models. In this paper, 
we investigate the vulnerability of single--image deepfake detectors to black--box attacks created by the newest generation of generative methods, namely Denoising Diffusion Models (DDMs). Our experiments are run on FaceForensics++, a widely used deepfake 
benchmark consisting of manipulated images generated with various techniques for face identity swapping and face reenactment. Attacks are crafted through guided reconstruction of existing deepfakes with a proposed DDM approach for face restoration. Our findings indicate that employing just a single denoising diffusion step in the reconstruction process of a deepfake can significantly reduce the likelihood of detection, all without introducing any perceptible image modifications. While training detectors using attack examples demonstrated some effectiveness, it was observed that discriminators trained on fully diffusion--based deepfakes exhibited limited generalizability when presented with our attacks.



\end{abstract}

\section{Introduction}\label{sec:intro}
With the rapid development of digital technologies, the generation of fake images and videos has become an almost effortless process. Although these methods have numerous benefits in the entertainment industry, they can as well be used for malicious purposes. Such examples are deepfakes, where the face of a target person is altered or used as a replacement for another person's face in order to fabricate certain scenarios~\cite{Rossler_FF+_2019}. The manipulated data can then be exploited to spread misinformation, harm the victim or manipulate public opinion. The development of accurate deepfake detection algorithms is therefore crucial for the prevention of possible violations.

\begin{figure}[!htb]
    \begin{center}
        \includegraphics[width=\linewidth]{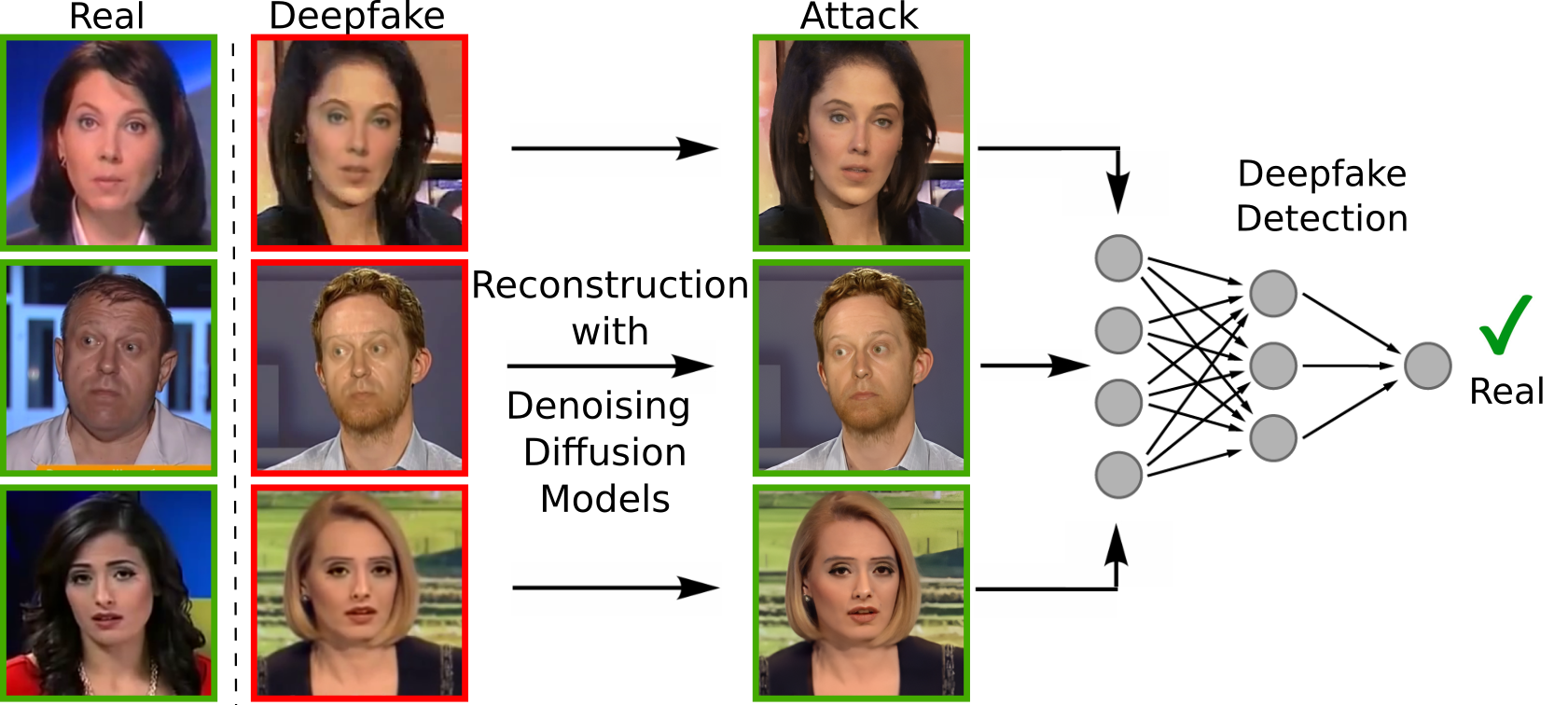}
        \caption{We investigate the vulnerability of popular single--image deepfake detection methods to \textbf{black--box attacks} generated by reconstructing existing deepfakes with Denoising Diffusion Models (DDMs). Our experimental results show that only one denoising step can generate an attack, that fouls the detectors into misclassifying the altered deepfake as real, all while preserving its visual appearance. Best viewed in color and zoomed in.} \label{fig:teaser}
    \end{center}
\vspace{-20pt}
\end{figure}

Over the years, various machine learning algorithms have been proposed for the automatic detection of manipulated data~\cite{Nguyen_multi_task_seg_2019, Luo_SRM_2021, Li_face_x_ray_2020}. These algorithms typically search for inconsistencies in lighting and shadows, visual artifacts, or unique fingerprints left by generative models during the creation of deepfakes. However, detection models can be vulnerable to specific attacks designed to intentionally deceive the detector, resulting in the misclassification of fake images as genuine. Traditional methods often generate such attacks by adding subtle perturbations to existing deepfake images~\cite{Hussain_adversarial_deepfakes_wacv_2021}. On the other hand, more sophisticated attacking techniques aim to integrate the attacks directly into the deepfake generation process, creating deepfakes that are more challenging 
in terms of their detectibility~\cite{carlini_latent_attacks_2020}.

The recent emergence Denoising Diffusion Models (DDMs), a new generation of generative models, has heightened concerns regarding the spread of fake data. These models have proven capable of generating fakes that are even more realistic and convincing than those produced by their predecessors, Generative Adversarial Networks (GANs)~\cite{Dhariwal_DDMs_beat_GANs_2021}. Initially developed for generating new data, DDMs have since found applications in a variety of other domains as well~\cite{diff_survey_pami_2023}. Recent studies have investigated the use of DDMs for manipulating facial expressions~\cite{zou_4D_Facial_Expression_2023}, performing face reenactment~\cite{stypulkowski_diffused_heads_2023}, generating morphing attacks~\cite{Damer_MorDIFF_2023}, and face swapping~\cite{kim_difface_swapping_2022, Zhao_diffswap_2023}.


Driven by the potential risks associated with this technology, 
we investigate the capability of DDMs to attack deepfake detection systems, by simply reconstructing existing deepfake images with a predetermined number of 
denoising steps (Figure~\ref{fig:teaser}). Generated black--box attacks are then used to test the accuracy of commonly used supervised and self--supervised detectors. In our study, we focus on deepfakes involving identity swaps and face reenactments. To the best of our knowledge, we are the first to investigate the potential exploitation of DDMs in this context.

In this paper, we make the following contributions: \textit{\textbf{i)}} We explore the ability of Denoising Diffusion Models (DDMs) to create black--box attacks targeting deepfake detection systems. \textit{\textbf{ii)}} We conduct a comprehensive assessment of the visual quality of the attacks. \textit{\textbf{iii)}} We evaluate the vulnerability of popular single--image deepfake detectors, to DDM intrusions. \textit{\textbf{iv)}} We analyze the detectability of our attacks when discriminators are trained on diffusion-based samples.


\begin{figure*}[!t]
    \begin{center}
        \includegraphics[width=\linewidth]{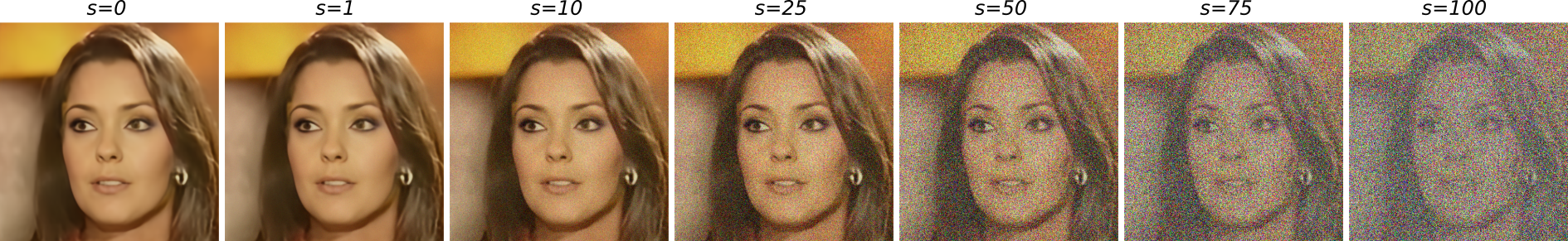}
        \caption{Visualisation of different \textbf{noise levels} applied to deepfake images, prior to their reconstruction with the proposed DDM. The variable $s$ refers to the number of applied noising steps. 
        The initial, unmodified deepfake is labeled with $s=0$. Best viewed in color.} \label{fig:noise_levels}
    \end{center}
\vspace{-15pt}
\end{figure*}

\section{Related work}
\noindent\textbf{Deepfake detection.} Recent single--image algorithms for deepfake detection are predominantly based on various deep learning methods. Naive approaches 
represent a CNN, that learns to differentiate between 
real and fake data. Xception~\cite{Chollet_Xception_2017} and MesoNet~\cite{Afchar_MesoNet_2018} are among the most popular in this category. To ensure the CNN has captured discriminative features, Nguyen~\textit{et al.} and Wang~\textit{et al.}~\cite{Nguyen_multi_task_seg_2019, Wang_RFM_2021_CVPR} both utilize explicit modeling of specific deepfake artifacts in the spatial space. Nguyen~\textit{et al.} also design Capsule~\cite{Nguyen_capsule_icassp_2019} to leverage the hierarchical and spatial relationships between image components. Recently, Cao~\textit{et al.} have proposed RECCE~\cite{Cao_RECCE_CVPR_2022}, an autoencoder that learns compact latent representations of real faces, hence classifies deepfakes as out--of--distribution samples with higher reconstruction error.

Yuyang~\textit{et al.}~\cite{Yuyang_F3Net_ECCV_2020} shift their focus to image analysis in the frequency space, recognizing that real and fake data typically have distinct frequency spectrums. They introduce F$^3$-Net, taking advantage of the frequency-aware image decomposition and local frequency statistics. Similarly, Liu~\textit{et al.}~\cite{Liu_SPSL_2021} note that unlike manipulated images, the phase spectrum of natural images preserves abundant frequency components. Their SPSL method combines this information with spatial clues, for a more robust deepfake detection. Another frequency--based approach SRM, proposed by Luo~\textit{et al.}~\cite{Luo_SRM_2021}, utilizes the high-frequency image noises and low--level RGB features, extracted by residual-guided spatial attention module.

Although very accurate when applied to a closed--set problem, supervised algorithms fail to generalize well to out--of--distrubution samples and images generated by unknown deepfake techniques. To address this problem, Li~\textit{et al.} present DSP-FWA~\cite{li_DSP-FWA_cvprw_2019}, a self--supervised method that does not rely on specific deepfake datasets but rather learns from simulated resolution inconsistencies in affine face warpings. In Face-X-Ray~\cite{Li_face_x_ray_2020} authors Li~\textit{et al.} take a similar approach with a focus on the blending artifacts of the deepfakes. Shiohara~\textit{et al.}~\cite{Shiohara_SBI_2022} further refine earlier approaches with a comprehensive simulation of common deepfake artifacts. Unlike previous simulated fake data, their Self--Blended Images (SBI) are created using a single face sample, that acts as both, source and target image.\\






\noindent\textbf{Attacks on deepfake detectors.}
Many studies have shown that deepfake detection methods can be prone to certain types of carefully crafted attacks. These attacks are generally divided into two main categories, i.e. white--box and black--box attacks. The former are designed with full knowledge about the architecture and the parameters of the targeted deepfake detector, while the latter involve trial and error to approximate the model under attack. 

Hussain~\textit{et al.}~\cite{Hussain_adversarial_deepfakes_wacv_2021} and Gandhi~\textit{et al.}~\cite{Gandhi_adversarial_fool_classifiers_ijcnn_2020} both conduct white--box attacks on popular naive deepfake detectors, such as Xception, MesoNet, VGG and ResNet. They optimize perturbations added to the deepfake images, in order to have this samples 
later classified as real. In the same context, Saminder~\textit{et al.}~\cite{Saminder_mitigating_adversarial_attacks_2023} explore various perturbation and AI techniques. Their findings suggest that incorporating adversarial samples during the training phase of a deepfake detector enhances both, its accuracy and robustness. However, it is important to note that while white--box attacks tend to be very efficient in constrained settings, they exhibit poor transferability, when applied to unknown systems. 

Differently from Saminder~\textit{et al.}~\cite{Saminder_mitigating_adversarial_attacks_2023}, Neekhara~\textit{et al.}~\cite{Neekhara_adversarial_threats_2021} evaluate various perturbation techniques in black--box settings.  In their study, they demonstrate that their generated adversarial samples have the capability to consistently attack different deepfake detection approaches.  Apart from image space perturbations, Carlini~\textit{et al.}~\cite{carlini_latent_attacks_2020} also investigate the implementation of perturbations in the latent space of the generative model, so that it yields adversarial images. 

A novel type of black--box attacks is introduced by Lou~\textit{et al.}~\cite{black_contrastive_Lou_2023}, who aim to mitigate the presence of GAN fingerprints, which are commonly used as clues in the deepfake detection process. To achieve this, they train an autoencoder that simultaneously learns to generate high--fidely images while applying imperceptible pixel perturbations. Conversely, Liu~\textit{et al.}~\cite{Liu_attacks_by_trace_removal_2023} perform blind post--processing of pre--generated deepfakes, by removing detectable traces left by the deepfake generation pipeline. The resulting deepfakes are therefore more authentic and challenging to detect. Similarly, Huang~\textit{et al.}~\cite{fake_polisher_huang_2020} develop FakePolisher, a shallow dictionary model, trained to accurately reconstruct only real data, thus efficiently removing typical GAN artifacts.  

Our work most closely resembles the last two studies~\cite{Liu_attacks_by_trace_removal_2023,fake_polisher_huang_2020}. In our approach, we leverage the capabilities of advanced Denoising Diffusion Models (DDMs) to perform guided post-processing of previously generated deepfakes. These processed images are then strategically utilized to conduct black-box attacks on both, supervised and self--supervised deepfake detection systems.

\section{Diffusion--based deepfake attacks}

In our study, we employ a DDM for conditional image synthesis. The architecture and the conditioning technique are based on guidelines published in the paper by Dhariwal~\textit{et al.}~\cite{Dhariwal_DDMs_beat_GANs_2021}. It's important to note that the model we use was not specifically designed nor optimized for attacking deepfake detection systems. Our main goal 
is to evaluate the effectiveness of this readily available diffusion--based model in generating black--box attacks. The objective of the generation process is to improve image quality of existing deepfakes and suppress detectable artifacts. 

\begin{figure}[!htb]
    \begin{center}
        \includegraphics[width=\linewidth]{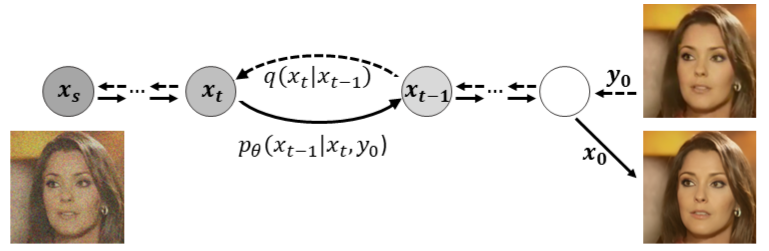}
        \caption{High--level overview of the \textbf{generation of deepfake attacks}. An existing deepfake image ($y_0$) is fed into the proposed diffusion--based model. The model creates a noisy sample $x_s$, by gradually adding Gaussian noise for $s$ steps. The deepfake attack ($x_0$) is then created by reversing the noising process.} \label{fig:model}
    \end{center}
\vspace{-15pt}
\end{figure}

The process of attack generation encompasses two stages. In the initial stage, referred to as the forward process, a selected deepfake image $y_0$ undergoes progressive corruption through the addition of Gaussian noise $\mathcal{N}(0, \sigma^2\mathbf{I})$, following a non-homogeneous Markov chain. Subsequently, in the reverse process stage, the deepfake image, now corrupted with $s$ noising steps ($x_s$), undergoes sequential denoising via a parametrized generative model $D_{\theta}(x,\sigma)$. In our experiments, $D_{\theta}$ represents a pretrained approximator, optimized to minimize the Kullback-Leibler (KL) divergence between the designed distribution $p(x_s|y_0)$ and its target distribution $q(x_s|x_0)$, with $x_0$ symbolizing the restored version of the initial input image $y_0$. A high--level illustration of the proposed deepfake attack generation process is depicted in Figure~\ref{fig:model}. Inspired by findings published in~\cite{Yue_difface_2022}, we treat the deepfake $y_0$ as a degraded version of its reconstruction $x_0$. We therefore model the marginal distribution of $x_s$ by implementing a diffused estimator, which guides the noising forward process.  


\begin{figure}[!hb]
    \begin{center}
        \includegraphics[width=\linewidth]{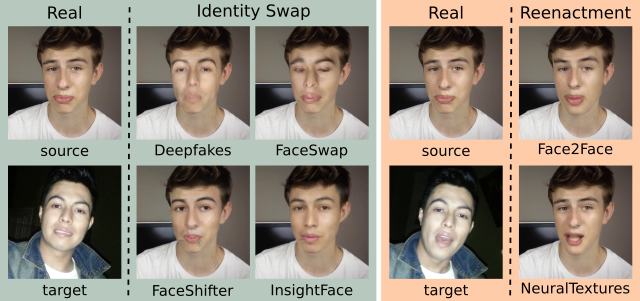}
        \caption{Our experiments are run on manipulated data sourced from \textbf{FaceForensics++} (FF++), where real faces are either reenacted, or the identity is swapped using a target image of another person. Manipulations include six different techniques: Deepfakes, FaceSwap, FaceShifter, InsightFace, Face2Face, and NeuralTextures, resulting in deepfakes of various qualities with diverse artifacts and fingerprints. Best viewed in color and zoomed in.} \label{fig:FF_data}
    \end{center}
\vspace{-15pt}
\end{figure}

\begin{table*}[t] 
\begin{center}
\resizebox{\linewidth}{!}{%
\begin{tabular}{|l ? c| c | c |c| c| c ? c| c | c |c| c| c ? c| c | c |c| c| c?} 
 \hline
 \textbf{Fake} & \multicolumn{6}{c?}{\textbf{SSIM \big\uparrow}}& \multicolumn{6}{c ?}{\textbf{LPIPS \big\downarrow}}& \multicolumn{6}{c ?}{\textbf{CSIM \big\uparrow}}\\ \cline{2-19}
  \textbf{data} & \textbf{s=1} & \textbf{s=10} & \textbf{s=25} & \textbf{s=50} & \textbf{s=75} & \textbf{s=100} & \textbf{s=1} & \textbf{s=10} & \textbf{s=25} & \textbf{s=50} & \textbf{s=75} & \textbf{s=100} & \textbf{s=1} & \textbf{s=10} & \textbf{s=25} & \textbf{s=50} & \textbf{s=75} & \textbf{s=100}\\ 
 \hline\hline
 DF & $0.9504$ & $0.9475$ & $0.9330$ & $0.9045$ & $0.8755$ & $0.8457$& $0.0626$ & $0.0625$ & $0.0714$ & $0.0951$ & $0.1145$ & $0.1286$ & $0.9593$ & $0.9366$ & $0.8577$ & $0.7015$ & $0.5500$ & $0.4008$\\ \hline
 F2F & $0.9528$ & $0.9496$ & $0.9349$ & $0.9066$ & $0.8782$ & $0.8496$& $0.0551$ & $0.0552$ & $0.0639$ & $0.0850$ & $0.1023$ & $0.1145$ & $0.9699$ & $0.9481$& $0.8733$ & $0.7210$ & $0.5665$ & $0.4105$\\ \hline 
 FSh & $0.9488$ & $0.9451$ & $0.9309$ & $0.9041$ & $0.8772$ & $0.8496$& $0.0659$ & $0.0660$ & $0.0742$ & $0.0960$ & $0.1138$ & $0.1265$ & $0.9712$ & $0.9482$ & $0.8732$ & $0.7153$ & $0.5490$ & $0.3895$\\ \hline 
 FS & $0.9544$ & $0.9507$ & $0.9352$ & $0.9048$ & $0.8744$ & $0.8435$ & $0.0513$ & $0.0519$ & $0.0597$ & $0.0811$ & $0.0994$ & $0.1129$ & $0.9687$ & $0.9479$ &$0.8777$ & $0.7295$ & $0.5709$ & $0.4127$\\ \hline  
 IF & $0.9457$ & $0.9432$ & $0.9302$ & $0.9042$ & $0.8775$ & $0.8500$ & $0.0722$ & $0.0716$ & $0.0802$ & $0.1038$ & $0.1229$ & $0.1363$ & $0.9725$ & $0.9534$ & $0.8834$ & $0.7275$ & $0.5629$ & $0.3981$\\ \hline
 NT & $0.9482$ & $0.9455$ & $0.9318$ & $0.9046$ & $0.8772$ & $0.8489$ & $0.0673$ & $0.0670$ & $0.0760$ & $0.0986$ & $0.1167$ & $0.1293$ & $0.9670$ & $0.9462$ & $0.8754$ & $0.7290$ & $0.5771$ & $0.4217$\\ \hline 
 \hline
  \textbf{Avg.} & $\mathbf{0.9505}$ & $\mathbf{0.9469}$ & $\mathbf{0.9327}$ & $\mathbf{0.9048}$ & $\mathbf{0.8767}$ & $\mathbf{0.8479}$ & $\mathbf{0.0624}$ & $\mathbf{0.0624}$ & $\mathbf{0.0709}$ & $\mathbf{0.0933}$ & $\mathbf{0.1116}$ & $\mathbf{0.1247}$ & $\mathbf{0.9681}$ & $\mathbf{0.9467}$ & $\mathbf{0.8735}$ & $\mathbf{0.7206}$ & $\mathbf{0.5627}$ & $\mathbf{0.4055}$\\
  \hline
\end{tabular}}
\end{center}
\caption{\textbf{Quantitative evaluation of generated attacks} using three different measures, i.e. Structural Similarity Index Measure (SSIM), Perceptual Image Patch Similarity (LPIPS) and  Cosine Similarity Index Measure (CSIM). Each attack is compared to its corresponding unmodified FF++ deepfake created with one of the 6 available techniques, here denoted by DF (Deepfakes), F2F (Face2Face), FSh (FaceShifter), FS (FaceSwap), IF (InsightFace) and NT (NeuralTextures). Generation of attacks with up to $s=50$ denoising steps does not significantly impact the image structure, while $s=75$ and $s=100$ alter the image to a point where the initial identity is degraded.}
\label{tab:visual_quality}
\end{table*}

\begin{figure*}[t]
\begin{center}
\centering
  \includegraphics[width=1\textwidth]{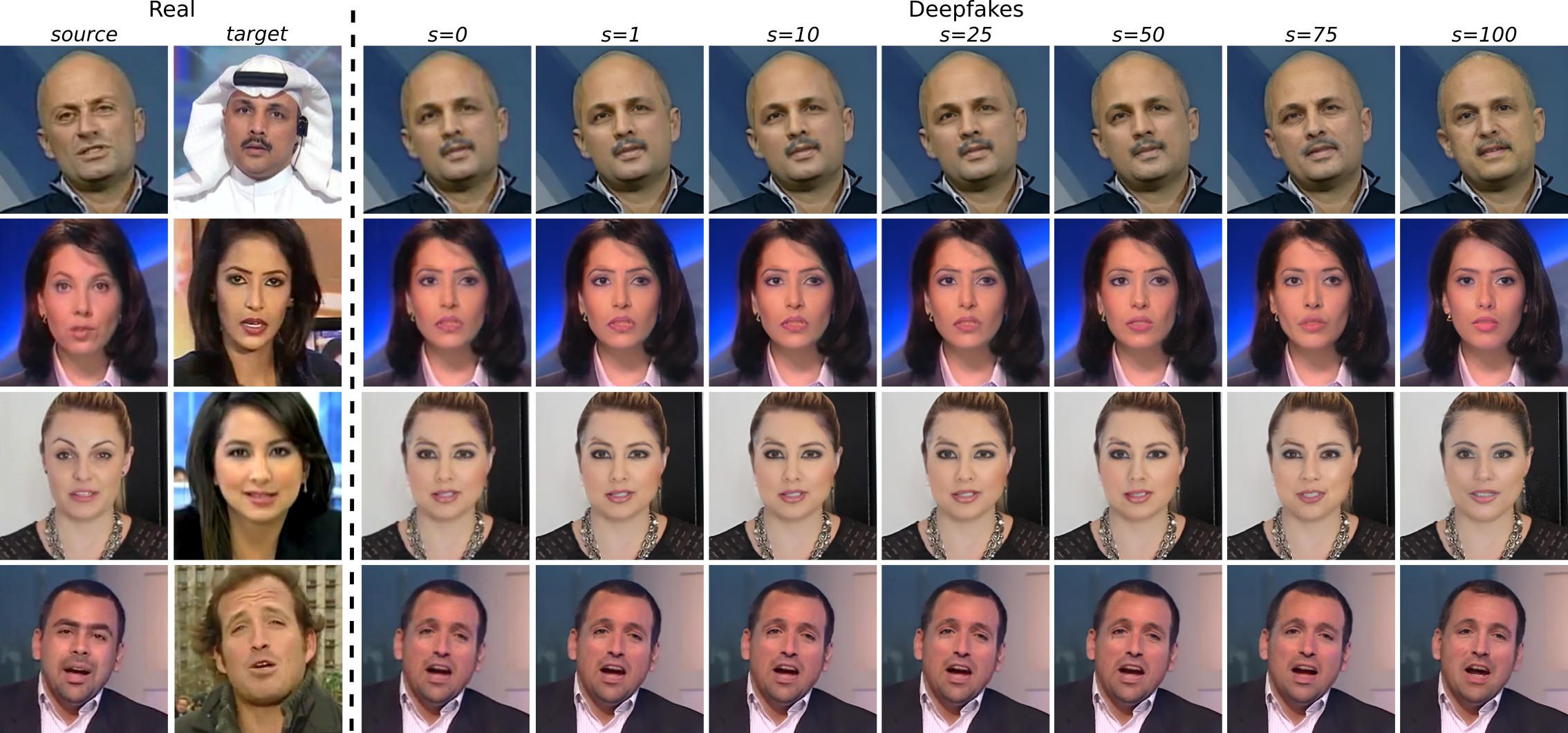}
\end{center}
\caption{\textbf{Qualitative comparison of generated attacks.} Low number $s$ of denoising diffusion steps introduces minimal to no visually perceptible modifications in the image structure. Higher $s$ values are on the other hand associated with notably enhanced image quality. The highest considered value $s=100$ seems to completely remove deepfake inconsistencies, but in some cases might change the appearance of different face parts. Unmodified FF++ deepfakes ($s=0$) are sourced from the FaceShifter subset. Best viewed in color and zoomed in. \label{fig:examples}}
\vspace{-5 pt}
\end{figure*}

\section{Experiments}\label{sec:experiments}

\noindent\textbf{Datasets.} In our study we experiment with FaceForensics++ (FF++)~\cite{Rossler_FF+_2019},  a benchmark commonly used for the evaluation of deepfake detection methods. The dataset consists of $1000$ real YouTube videos, each available in three different qualities. We only use the highest--quality, namely the raw data. Originally, FF++ deepfakes were generated using two face-swapping techniques, i.e. Deepfakes and FaceSwap, and two face reenactment methods, i.e. Face2Face and NeuralTextures. As GAN-based deepfake quality has improved over time, FaceShifter deepfakes have been later included in the dataset. This advanced face-swapping technique not only yields higher-quality fake data but also handles occlusions more effectively than competitive methods. In this study, we additionally consider InsightFace, a popular and widely used face swapping model for generation of high--fidelity deepfakes. We use the implementation of this method provided by the open--source GitHub~\cite{insightface} toolbox. The inference with the model is performed using InsightFace weights pretrained on face crops of size $128\times128$ px. Examples of FF++ samples generated with each of the aforementioned manipulation methods are presented in Figure~\ref{fig:FF_data}. As can be seen from the Figure, individual methods produce different image qualities in terms of fidelity, and presence of artifacts.

In this paper, we focus on single-image deepfake detectors and therefore opt to extract only every $10$th frame from each real and fake video sequence. The detection and cropping of facial regions are conducted using a pretrained MTCNN model~\cite{Xiang_MTCNN_2017}. All extracted deepfakes undergo reconstruction using our proposed DDM approach, applied six times 
with varying numbers of diffusion steps $s$: $1$, $10$, $25$, $50$, $75$, and $100$. Noise levels representing individual $s$ values are visualized in Figure~\ref{fig:noise_levels}. In our experiments, we adhere to the FF++ train, validation and test split.




\noindent\textbf{Experimental details.} Deepfake attacks are created utilizing DDM weights that are pretrained on the FFHQ dataset~\cite{Karras_progan_FFHQ_iclr_2018}. SwinIR~\footnote{https://github.com/zsyOAOA/DifFace}, a transformer commonly used in image restoration tasks, serves as a backbone of the diffused estimator. These generated attacks are then leveraged to assess the vulnerability of $9$ widely used deepfake detection models. We evaluate Xception~\cite{Chollet_Xception_2017}, MesoInception~\cite{Afchar_MesoNet_2018}, Capsule~\cite{Nguyen_capsule_icassp_2019} and RECCE~\cite{Cao_RECCE_CVPR_2022} as representatives of discriminative approaches based on spatial features. Among frequency--based detectors we consider F$^3$-Net~\cite{Yuyang_F3Net_ECCV_2020} and SRM~\cite{Luo_SRM_2021}. All these detectors are trained from scratch, in a supervised manner. In the testing phase, we use the weights of the model that achieves best AUC score on the validation image subset. Independent models are trained for each of the $6$ FF++ deepfake methods.
The vulnerability of self--supervised methods, is evaluated using $3$ models, i.e. DSP-FWA~\cite{li_DSP-FWA_cvprw_2019}, Face X-Ray~\cite{Li_face_x_ray_2020}, and Self Blended Images (SBI)~\cite{Shiohara_SBI_2022}.  The training of these detectors is performed using real FF++ data only. To compensate for the missing class of deepfakes, we use augmented images (simulated fakes), as suggested in corresponding papers. Experiments were run on NVIDIA GeForce RTX 3090.

\noindent\textbf{Evaluation metrics.} The visual quality of generated attacks is measured by comparing them to corresponding unmodified deepfakes. Their perceived quality is estimated with Structural Similarity Index Measure (SSIM). For comparison of higher--level image features we use Learned Perceptual Image Patch Similarity (LPIPS), based on a pretrained SqueezeNet~\cite{SqueezeNet_ICLR_2017}. Finally, the preservation of the identity embedded in the unmodified deepfake is calculated with Cosine Similarity Index Measure (CSIM). Identity vectors are extracted by AdaFace~\cite{kim_adaface_2022}.

Deepfake detectors, once trained, are assessed separately on standard FF++ deepfakes and on attacks generated by reconstructing original manipulated data with various noise levels $s$. To ensure \textit{fair comparison} across individual assessment runs and to simulate a \textit{real--world scenario}, each detection method is first evaluated on the validation subset of real and \textit{unmodified} deepfake images. Calculated threshold at the Equal Error Rate (EER) point on the ROC is then used for the classification of testing deepfake samples and as well as attacks. The performance of the detectors is measured in terms of True Positive Rate (TPR), which denotes the proportion of accurately identified fake data. 

\section{Results}
In this section, we first assess the quality of generated diffusion--based attacks. Next, we evaluate the vulnerability of deepfake detectors to these attacks. Additionaly, we investigate whether the accuracy of discriminative detectors improves, when we train them using the attacks as fake samples. Finally, we investigate the vulnerability of discriminative detectors trained with fully diffusion-based face swaps. 

\begin{figure}[!ht]
    \begin{center}
        \includegraphics[width=\linewidth]{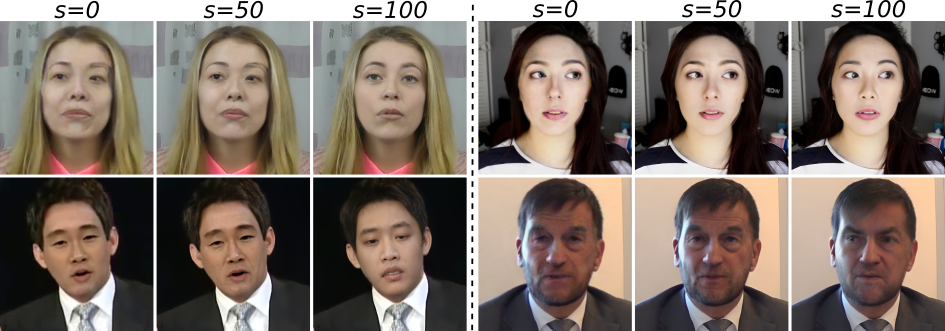}
        \caption{
        \textbf{Severe examples of image alterations} resulting from the reconstruction of original deepfakes ($s=0$) with a high number of denoising steps ($s=100$). Unlike $s=50$ examples, samples generated with $s=100$ include significant modifications in soft biometric attributes like age, hairstyle, race, and ethnicity. We also observe subtle changes in the shape and color of the eyes, nose, and lips. Best viewed in color and zoomed in.} \label{fig:failure}
    \end{center}
\vspace{-15pt}
\end{figure}

\noindent\textbf{Quality assessment of deepfake attacks.} The generation of deepfake attacks with the proposed DDM approach introduces inevitable image changes. The quantitative assessment of their visual quality is given in Table~\ref{tab:visual_quality}. Calculated SSIM and LPIPS values suggest that reconstructing deepfakes with up to $s=50$ denoising steps does not significantly change their structure. A higher value of $s$ on the other hand visibly modifies the initial image. These modifications decrease the CSIM value, indicating that the identity of the face has been degraded to some extent. 

These findings are also supported by the qualitative analysis of attacks. As can be seen in Figure~\ref{fig:examples}, there are no obvious, perceivable \emph{structural} differences between unmodified deepfakes ($s=0$) and attacks denoted by $s=1$, $s=10$, $s=25$ and $s=50$. We observe that higher levels of noise in general produce more realistic images, in terms of fidelity. Moreover, when s=100, the DDM is capable of completely removing typical deepfake inconsistencies. In the 2nd and 3rd row of Figure~\ref{fig:examples} for instance, the DDM has successfully removed double eyebrows. In the 2nd row, the irregularities in the teeth shape and position are also fixed, while unnatural forehead shadows are eliminated. This noise level, however, often modifies the appearance (size, shape, color) of individual facial parts. In the 1st row of Figure~\ref{fig:examples} for example, we can see how the DDM has added hair to the initially bald head. Similar changes can be seen in the 4th row, where white hair strands have been inserted. In some cases, subtle changes in the microexpression of the deepfakes can also be observed. Some severe examples, where the application of a high number of diffusion denoising steps ($s=100$) has drastically changed the initial face are shown in Figure~\ref{fig:failure}. Here, we can see evident differences in soft biometric attributes such as age, hairstyle, race and ethnicity. We can also notice subtle changes in the eyes, nose and lip shape and color. 

\begin{figure*}[!htb]
    \begin{center}
        \includegraphics[width=0.97\linewidth]{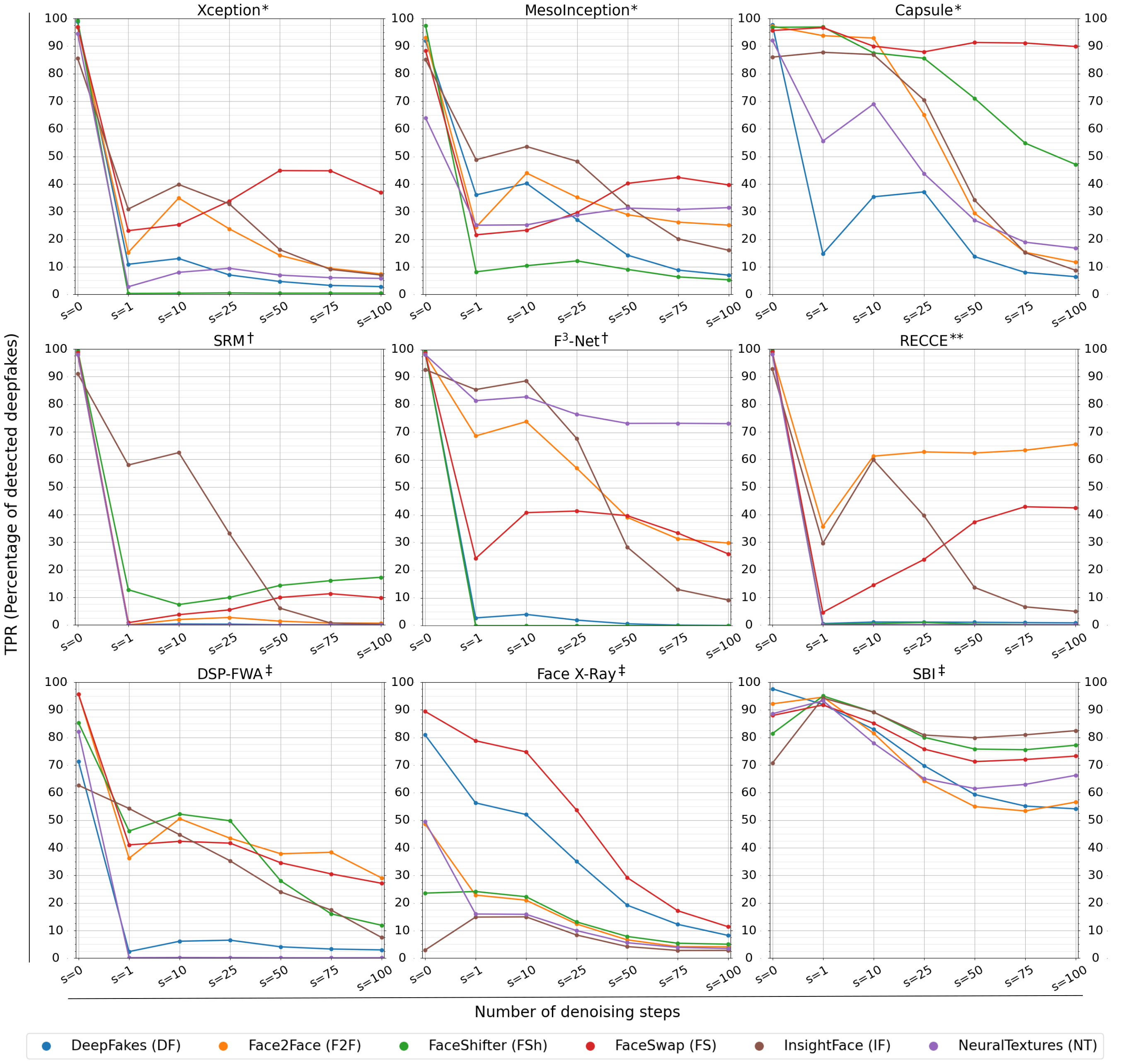}
        \caption{Percentage of \textbf{detected FF++ deepfakes ($\mathbf{s=0}$) and attacks ($\mathbf{s>0}$)}. The threshold used for the classification of manipulated images is calculated at the EER point of the ROC evaluated on real and unmodified fake images ($s=0$) from the validation subset. Types of evaluated detectors: ($\ast$) discriminator based on spatial features, ($\dag$) discriminator based on frequency features, ($\ast\ast$) autoencoder based on reconstruction error, ($\ddag$) self-supervised method. We observe, that self--supervised methods (see SBI) are in general more robust to our attacks than discriminative models. However, all considered models, irrespective of their type, experience some degree of adverse impact.} \label{fig:TPR}
    \end{center}
\vspace{-15pt}
\end{figure*}

\begin{table*}[!t] 
\begin{center}
\resizebox{\linewidth}{!}{%
\begin{tabular}{|l ? c| c | c ?c| c| c ? c| c | c ? c| c| c ? c| c | c ? c| c| c?} 
\cline{2-19}
\multicolumn{1}{c?}{} & \multicolumn{18}{c?}{\textbf{train:} \textit{real, unmodified deepfakes, attacks}} \\ \cline{2-19}
 \multicolumn{1}{c?}{} & \multicolumn{3}{c?}{\textbf{Deepfakes (DF)}} & \multicolumn{3}{c ?}{\textbf{Face2Face (F2F)}} & \multicolumn{3}{c?}{\textbf{FaceShifter (FSh)}} & \multicolumn{3}{c?}{\textbf{FaceSwap (FS)}} & \multicolumn{3}{c ?}{\textbf{InsightFace (IF)}} & \multicolumn{3}{c ?}{\textbf{NeuralTextures (NT)}}\\ \cline{1-19}
  \multicolumn{1}{|c?}{\textbf{test}} & \textbf{s=1} & \textbf{s=50} & \textbf{s=100} & \textbf{s=1} & \textbf{s=50} & \textbf{s=100} & \textbf{s=1} & \textbf{s=50} & \textbf{s=100} & \textbf{s=1} & \textbf{s=50} & \textbf{s=100} & \textbf{s=1} & \textbf{s=50} & \textbf{s=100} & \textbf{s=1} & \textbf{s=50} & \textbf{s=100}\\ 
 \hline
 \textbf{s=0} & $76.02$ & $78.74$ & $84.16$ & $91.16$ & $90.11$ & $95.33$ & $89.17$ & $94.44$ & $85.27$ & $92.14$ & $91.35$ & $93.04$ & $77.74$ & $77.69$ & $74.31$ & $73.99$ & $86.25$ & $70.18$\\ \hline
 \textbf{s=1} & $\mathbf{99.31}$ & $\mathbf{96.06}$ & $\mathbf{99.27}$ & $\mathbf{99.19}$ & $\mathbf{96.80}$ & $\mathbf{99.57}$& $\mathbf{96.21}$ & $\mathbf{96.29}$ & $87.21$ & $\mathbf{96.39}$ & $96.95$ & $87.43$ & $\mathbf{99.92}$ & $\mathbf{93.66}$ & $88.36$ & $\mathbf{94.17}$ & $99.30$ & $64.73$\\ \hline
 \textbf{s=10} & $45.64$ & $94.21$ & $87.27$ & $64.37$ & $95.03$ & $80.04$& $89.38$ & $94.03$ & $76.94$ & $85.11$ & $87.04$ & $52.39$ & $73.21$ & $92.78$ & $69.09$ & $92.38$ & $97.85$ & $5.80$\\ \hline 
 \textbf{s=25} & $37.28$ & $92.87$ & $88.42$ & $58.17$ & $94.06$ & $83.80$& $84.65$ & $94.14$ & $82.96$ & $87.57$ & $91.51$ & $69.96$ & $66.80$ & $93.10$ & $82.77$ & $83.76$ & $98.51$ & $23.10$\\ \hline 
 \textbf{s=50} & $36.19$ & $93.64$ & $89.55$ & $63.43$ & $95.46$ & $95.75$& $71.73$ & $95.11$ & $91.88$ & $93.00$ & $97.28$ & $92.21$ & $67.24$ & $93.21$ & $96.35$ & $61.09$ & $99.40$ & $67.98$\\ \hline  
 \textbf{s=75} & $31.68$ & $89.87$ & $96.49$ & $63.43$ & $95.27$ & $98.25$& $61.33$ & $94.46$ & $95.59$ & $93.73$ & $\mathbf{98.21}$ & $97.21$ & $67.54$ & $92.70$ & $98.72$ & $46.64$ & $99.70$ & $86.21$\\ \hline
 \textbf{s=100} & $29.14$ & $86.24$ & $98.78$ & $60.69$ & $92.85$ & $98.95$& $57.05$ & $93.36$ & $\mathbf{96.45}$ & $92.90$ & $98.11$ & $\mathbf{97.51}$ & $67.43$ & $92.59$ & $\mathbf{99.16}$ & $42.86$ & $\mathbf{99.97}$ & $\mathbf{91.28}$\\ \hline
\hline \textbf{Avg.} & $50.75$ & $90.23$ & $\mathbf{93.63}$ & $71.49$ & $\mathbf{94.23}$ & $93.10$ & $78.50$& $\mathbf{94.55}$ & $88.04$ & $91.55$ & $\mathbf{94.35}$ & $84.25$ & $74.26$ & $\mathbf{90.82}$ & $86.96$ & $70.70$ & $\mathbf{97.28}$ & $58.47$ \\ \hline
\end{tabular}}
\end{center}
\caption{TPR results indicating the \textbf{detectability of black-box attacks}, when Xception is discriminatively trained on a dataset that consists of real samples, unmodified deepfakes, and either $s=1$, $s=50$, or $s=100$ attacks. While training the detector on $s=1$ attacks gives poor overall results, training on $s=100$ significantly improves the detection accuracy of all types of attacks. Training the discriminator on $s=50$ yields the most consistent and effective detection results across different attacks, with a TPR between $90\%$ and $97\%$.}
\label{tab:detectibility}
\end{table*}

\begin{figure*}[t]
\begin{center}
\centering
  \includegraphics[width=1\textwidth]{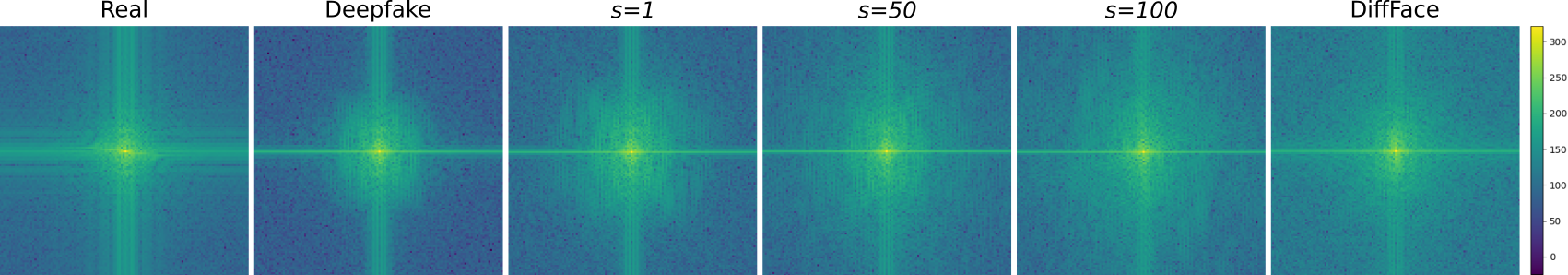}
\end{center}
\caption{Visualisation of the \textbf{Discrete Fourier Transform (DFT) spectra} of real data, unmodified deepfakes, our DDM attacks ($s=1$, $s=50$, $s=100$) and diffusion--based face swapping images generated with DiffFace~\cite{kim_difface_swapping_2022}. Our proposed DDM--based approach for the generation of attacks reintroduces high-frequency elements in the low--density spectral range of deepfakes, mimicking the spectrum of real images. Entirely diffusion--based methods for creation of deepfake data (such as DiffFace) in general tend to overestimate the frequency density of real data. These spectral discrepancies can be utilized, to expose the DDM deepfakes. Best viewed in color and zoomed in.\label{fig:dft}}
\vspace{-15pt}
\end{figure*}

\noindent \textbf{Vulnerability of deepfake detectors to attacks.} We evaluate considered deepfake methods following the protocol described in Section~\ref{sec:experiments} (see \textit{Evaluation metrics}). Figure~\ref{fig:TPR} depicts obtained results in terms of True Positive Rate (TPR). We observe, that attacks generated with only one DDM step ($s=1$) can severely affect the accuracy of detectors. Discriminative methods tend to be more prone to DDM attacks as opposed to self--supervised methods. Capsule is the least vulnerable among discriminative detectors. However, it experiences a severe drop in accuracy when $s>25$. Naive spatial classifiers Xception and MesoInception, along with the frequency--based classifier SRM show overall worst results in terms of the percentage of detected attacks. These methods maintain a consistent TPR of above $90\%$ across different unmodified deepfakes, but the TPR of reconstructed deepfakes is in most cases far below $50\%$. 
RECCE exhibits similar problems, except when analyzing Face2Face reenacted faces and FaceSwap samples. When these fake images are processed with $s>1$ denoising steps, the reconstruction error of RECCE is closer to the error of unmodified fakes.  In general, higher number of denoising steps $s$ generates more challenging attacks, leading to lower TPR. In some cases, we observe a local peak in discriminative method's TPR calculated for $s=10$. We hypothesize, that this particular amount of denoising steps induces image artifacts similar to those present in unmodified deepfakes.

Unlike discriminative approaches, self--supervised models are trained on simulated fake data. Their performance is therefore strongly dependent on the simulation technique. Based on Figure~\ref{fig:TPR}, DSP-FWA, one of the very first self--supervised models, experiences a huge drop in accuracy (at least $30\%$ or more), when attacks with $s=1$ are presented. In contrast, Face X-Ray, a model trained on manipulated data that mimics deepfakes in a more sophisticated way, has slightly more stable performance on $s=1$ and $s=10$ attacks. This method is however very sensitive to modified deepfakes generated with $s>10$. Moreover, when Face X-Ray is presented with $s=100$ attacks, it detects less than $10\%$ of them. In our study, the most robust detector is SBI. Although its performance starts dropping gradually when $s>1$, the TPR in all tested scenarios stays above $50\%$. Interestingly, SBI initially detects InsightFace deepfakes with a TPR of around $70\%$, but corresponding attacks are recognized with a higher TPR of $80\%$ or more. We hypothesize that InsightFace traces are not well approximated by the SBI augmentation approach, whereas our DDM introduces frequencies that better resemble modeled artifacts.


\noindent \textbf{Detectability of deepfake attacks.} To investigate how detectable our black-box attacks are in discriminative settings, we conduct a study in which deepfakes reconstructed using our proposed DDM method are incorporated into the training data. Specifically, we train Xception on $3$ separate datasets, each containing only one type of attacks. The first set consists of real, unmodified fake data and attacks processed with $s=1$ diffusion step. In the second and the third training sets, $s=1$ attacks are replaced by $s=50$ and $s=100$ attacks, respectively. We adhere to the same evaluation protocol as before, this time incorporating corresponding validation attacks, as well. Obtained results are presented in Table~\ref{tab:detectibility}. Our empirical study reveals that, on average, $s=1$ attacks are the least challenging to detect, regardless of the type of attacks included in the training set. However, when Xception is trained on $s=1$ reconstructions, it performs poorly in the detection of other attacks. On the contrary, training on $s=100$ reconstructions greatly improves the detection accuracy of other attacks, as well. Even so, attacks with smaller $s$ values are detected much less accurately, than attacks with $s$ value closer to $100$. Training Xception on $s=50$ attacks achieves the best average results, across different FF++ Deepfake methods. Nonetheless, this discriminator fails to detect between $3\%$ to $10\%$ of deepfakes, depending on the training dataset. The best performance (TPR of $97.28\%$) is gained in experiments performed on NeuralTextures data, while the worst performance ($90.82\%$) is measured on the InsightFace data. 

To gain deeper insights into the differences between real, deepfake images and DDM-generated attacks, we analyze the data in the frequency domain. The magnitudes of the Discrete Fourier Transform (DFT) for these images are visualized in Figure~\ref{fig:dft}. We notice that deepfakes, in contrast to real images, exhibit a significantly lower spectrum density moving away from the center of the spectrum, which represents dominant low-frequency components. 
Our DDM approach for the generation of attacks aims to address this deficiency by reintroducing high-frequency elements. Specifically, by reconstructing an existing deepfake with $s=1$ diffusion step, we start compensating for the missing spectral components. As we increase the number of diffusion steps $s$, we intensify the higher spectral range. In order to investigate the differences between our deepfake attacks, and fake data generated by a DDM from scratch, in Figure~\ref{fig:dft} we also visualize the DFT spectrum of DiffFace~\cite{kim_difface_swapping_2022}, a diffusion model for face swapping.  We note that, unlike our proposed DDM approach, DiffFace overestimates the frequency density, causing spectral discrepancies, that can be utilized to expose deepfakes. Our findings are consistent with observations 
in~\cite{Ricker_general_deepfakes_detection_2022}, where similar results were obtained for other common diffusion models, as well. However, we note that neither our method nor DiffFace create strong spectral artifacts such as regular grids, typical for GAN--based models.

\noindent\textbf{Detecting DDM data and knowledge transfer to attacks.} In our final experiment, we evaluate the ability of a naive discriminator to identify pure diffusion deepfakes and explore its capacity for knowledge transfer, by attacking it with our diffusion--based attacks. For this purpose, we discriminatively train Xception on real, unmodified deepfakes and DiffFace face swaps. Results, obtained by following the previously established protocol, are summarized in Table~\ref{tab:diff_transferability}. We note that DiffFace deepfakes, which have a True Positive Rate (TPR) of over $99\%$, present a lesser challenge for the detector compared to all other non-diffusion Deepfake methods from FF++. Nevertheless, the TPR for these methods still exceeds $93\%$. InsightFace deepfakes are in this experiment the most challenging to detect, whereas the best detection results are achieved on FaceShifter (FSh), Face2Face (F2F), and Deepfakes (DF). Interestingly, despite the discriminator being trained on DiffFace diffusion samples, which in part share similar frequency spectrum with our reconstructed fakes (as illustrated in Figure~\ref{fig:dft}), it remains highly vulnerable to all three types of attacks considered in this comparison. Higher number of diffusion steps $s$ is in general associated with higher TPR. Nevertheless, the average TPR of attacks is relatively low. Lowest TPR (less than $10\%$) was measured on FaceShifter and NeuralTextures. Moderately low TPR (between $25$ and $45\%$) was measured on Deepfakes, Face2Face and InsightFace, while highest TPR was calculated on FaceSwap $s=50$ and $s=100$ attacks ($66.51\%$ and $81.30\%$, respectively). These isolated peaks in the TPR however, can not be entirely attributed to the training with DiffFace samples, as we observe the same trend in experiments, where we do not use any diffusion-based train images (see Figure~\ref{fig:TPR}).

 

\begin{table}[!t] 
\begin{center}
\resizebox{0.99\columnwidth}{!}{%
\begin{tabular}{|l | r | r | r | r | r | r|} 
\cline{2-7}
\multicolumn{1}{c|}{} & \multicolumn{6}{c|}{\textbf{train:} \textit{real, unmodified deepfakes, DiffFace face swaps}} \\\hline
 \multicolumn{1}{|c|}{\textbf{test}} & \multicolumn{1}{c|}{\textbf{DF}} & \multicolumn{1}{c |}{\textbf{F2F}} & \multicolumn{1}{c|}{\textbf{FSh}} & \multicolumn{1}{c|}{\textbf{FS}} & \multicolumn{1}{c |}{\textbf{IF}} & \multicolumn{1}{c|}{\textbf{NT}}\\ \hline
 \textbf{DiffFace}~\cite{kim_difface_swapping_2022} 
              & $100.00$ & $100.00$ & $99.44$ & $99.76$ & $100.00$ & $100.00$ \\ \hline
 \textbf{s=0} & $98.54$ & $98.33$ & $98.98$ & $95.26$ & $93.19$ & $96.69$ \\ \hline
 \textbf{s=1} & $26.60$ & $30.33$ & $8.47$ & $34.48$ & $32.63$ & $5.80$ \\ \hline
 \textbf{s=50} & $29.28$ & $30.71$ & $3.23$ & $66.51$ & $37.18$ & $6.36$\\ \hline
 \textbf{s=100} & $39.17$ & $26.87$ & $2.23$ & $81.30$ & $44.44$ & $8.39$\\ \hline
\hline \textbf{Avg.} & $\mathbf{58.72}$ & $\mathbf{57.25}$ & $\mathbf{42.47}$ & $\mathbf{75.46}$ & $\mathbf{61.49}$ & $\mathbf{43.45}$ \\ \hline
\end{tabular}}
\end{center}
\caption{TPR results achieved by Xception, when it is discriminatevly trained on real images, different FF++ face manipulations and DiffFace (DDM) deepfakes. Measured accuracy  indicates the \textbf{detectibility of DDM data and the generalization} capacities of the discriminator. While purely diffusion--based face swaps DiffFace are detected with a TPR of over $99\%$, the discriminator is still highly vulnerable to our diffusion--based attacks.}
\label{tab:diff_transferability}
\end{table}

\section{Conclusion}
Recently discovered Denoising Diffusion Models (DDMs) have shown impressive capabilities for generating highly realistic and convincing images. In this paper, we investigate their potential employment as generators of black--box attacks on deepfake detection systems. We utilize a guided conditional DDM to reconstruct FaceForensics++ (FF++) deepfakes with a predetermined number of diffusion steps, which was found to directly relate to the properties of the generated attack. The proposed DDM approach not only improves the visual quality of an existing deepfake, but it also removes typical deepfake artifacts. 

We leverage generated black--box attacks in the evaluation of the vulnerability of commonly used deepfake detectors. Our study reveals that reconstructing deepfakes with just one diffusion step can significantly decrease the accuracy of the deepfake detectors. A higher number of steps generally leads to lower detection accuracy. Self--supervised detectors tend to be more robust than discriminative spatial or frequency--based methods. Training a naive discriminator using attacks generated with a specific number of denoising steps effectively identifies manipulations associated with that particular level of diffusion noise. However, this approach lacks generalizability to attacks produced with a different number of denoising steps. We also investigate the vulnerability of the discriminator that has been trained using purely diffusion-based face manipulation. Although this model is slightly less vulnerable to our DDM attacks in comparison to models trained on non--diffusion samples, it still shows limited generalizability.





{\small
\bibliographystyle{ieee_fullname}
\bibliography{egbib}
}

\end{document}